\documentclass[preprint,12pt]{elsarticle}

\usepackage{amssymb}

\journal{Pattern Recognition Letters}

\begin{document}

\begin{frontmatter}



\title{Towards End-to-End Neural Face Authentication in the Wild – Quantifying and Compensating for Directional Lighting Effects.}


\author[inst1]{Viktor Varkarakis}

\affiliation[inst1]{organization={Department of Electronic Engineering, College of Engineering, National University of Ireland Galway},
            addressline={University Road}, 
            city={Galway},
            country={Ireland}}

\author[inst1]{Wang Yao}
\author[inst1]{Peter Corcoran}

\begin{abstract}
The recent availability of low-power neural accelerator hardware, combined with improvements in end-to-end neural facial recognition algorithms provides enabling technology for on-device facial authentication. The present research work examines the effects of directional lighting on a State-of-Art (SoA) neural face recognizer. A synthetic re-lighting technique is used to augment data samples due to the lack of public data-sets with sufficient directional lighting variations. Top lighting and its variants (top-left, top-right) are found to have minimal effect on accuracy, while bottom-left or bottom-right directional lighting have the most pronounced effects. Following the fine-tuning of network weights, the face recognition model is shown to achieve close to the original Receiver Operating Characteristic curve (ROC) performance across all lighting conditions, and demonstrates an ability to generalize beyond the lighting augmentations used in the fine-tuning data-set. This work shows that a SoA neural face recognition models can be tuned to compensate for directional lighting effects, removing the need for a pre-processing step prior to applying facial recognition. 

\end{abstract}


\end{frontmatter}


\section{Introduction}
\label{sec1}

Human Face Recognition (FR) has been an active research field in computer vision since the early 1990’s \citep{Turk1991Eigenfaces} and early Convolutional Neural Network (CNN) based approaches were in evidence before the end of that decade \citep{Lawrence1997Face}. Over the last two decades FR has been well-studied in the literature with the most recent advances being driven by advances in CNN and deep learning \citep{Schroff2015Facenet:,Taigman2014Deepface:,Deng2019Arcface:,Wang2018Cosface:,mehdipour2016comprehensive}. In much of the literature the test samples for FR are assumed to be normalized in terms of pose, facial expression and illumination to simplify the challenge of accurately distinguishing an individual identity among a very large population. But, as it is not always feasible to capture optimal facial samples, some studies have explored the effects of different factors on the accuracy of State-of-Art (SoA) FR systems. 

The main factors that affect FR include (i) pose \citep{Tran2017Disentangled,Zhao2018Towards,Choi2011Face,Lanitis1997Automatic}, (ii) illumination \citep{Beveridge2010Quantifying,Zhu2017Illumination,Wang2018Illumination}, (iii) facial expression \citep{pena2020facial,Peng2018Facial,Pavlov2018Facial}, (iv) age \citep{Riaz2019Age-invariant,Deb2017Face} and, (v) gender variations \citep{Werther2018Gender,Narang2016Gender}.
  
In this work, the focus is on the latest end-to-end fully neural FR techniques \citep{Du2020Elements} as these represent current SoA in terms of accuracy and have the potential for implementation in the latest neural accelerators \citep{Corcoran2019Deep,Goel2020Survey}. The initial focus for implementation of neural algorithms in embedded devices was on network optimizations such as parameter quantization and pruning, compressed convolutional filters and matrix factorizations \citep{Goel2020Survey}. However, the attention has recently shifted towards specialized neural topologies \citep{Li2016Ternary,He2018Optimize} and ultra-low power realizations in hardware \citep{Fleischer2018Scalable,Guicquero2020Algorithmic}. Such optimizations enable a SoA neural FR architectures to be implemented in a low-power consumer appliance, enabling a new generation of devices capable of identifying their owners, provide access control and personalize the device’s responses and behaviour. However, this introduces new challenges as such FR embodiment can no longer rely on pre-processing of input facial samples to optimize power consumption. Thus, all image processing must be achieved in a fully neural implementation, requiring a neural FR to be robust to factors such as pose and illumination. Here our goal is to determine the feasibility of modifying a high accuracy SoA neural FR architecture to demonstrate robustness to uncompensated input image samples. \\

This leads to the research questions posed in this work: (i) can we better quantify the effects of the external factors that affect fully neural FR and develop metrics to evaluate these; and (ii) can a fully neural FR architecture be modified through tuning and/or re-training to compensate internally for such external factors? As human interaction with a consumer device will typically lead the user to look directly at a screen, or similar user-interface, this work has a focus on the effects of lighting/illumination scenarios on FR. A consideration of the effects of facial pose and other external factors is left to future studies. Specifically for the lighting variation, numerous image pre-processing methods exist to improve the performance of the FR model \citep{ruiz2008illumination,Zou2007Illumination,Shinwari2019Comparative}, but studies exploring the tuning or training of FR models to be robust to lighting variations are relatively rare \citep{crispell2017dataset,le2019illumination}.\\ 

As a first step towards answering these research questions, this work employs a SoA re-lighting methodology to augment a set of high-quality facial images with directional lighting effects. The effect of these augmentations on the performance of a SoA FR method is quantified using Receiver Operating Characteristic curve (ROC) techniques. A similar approach was used recently to validate synthetic facial identities \citep{varkarakis2020validating}. Note that a re-lighting augmentation approach was adopted as existing public datasets do not provide sufficient lighting variability. This is discussed in the sections \ref{effect of light} and \ref{conclusion and future work}.  Finally, this works studies the feasibility of handling lighting variations by fine-tuning the neural FR network. The results for directional lighting are promising and indicate the potential for an end-to-end neural face authentication solution for in-the-wild faces. 

\section{Related Works}
\label{related works}

The advances in computational resources and with a surge in access to very large datasets, deep learning architectures have been developed and pushed the SoA in the FR task achieving exceptional accuracy results \citep{mehdipour2016comprehensive}. Famous deep neural approaches include DeepFace, FaceNet and ArcFace \citep{Schroff2015Facenet:,Taigman2014Deepface:,Deng2019Arcface:,Wang2018Cosface:}. Each work advanced the accuracy on the benchmarks of FR and new loss functions, pre-processing techniques and deep neural architectures were introduced. More information regarding the SoA of deep neural FR approaches as well as the entire pipeline of the FR and the methods used are given in \citep{Du2020Elements,adjabi2020past}.  

Despite the improvements, the FR task remained challenging in several cases. Studies revealed that many factors can have a negative impact on the FR performance, with the main factors being pose, illumination and others \citep{Beveridge2010Quantifying,mehdipour2016comprehensive,Lanitis1997Automatic,Mahmood2016Effects}. 
Specifically, on face samples with lighting variation, techniques were proposed that compliment both the traditional and deep learning FR methods reporting improved performance \citep{mehdipour2016comprehensive,Han2013comparative}. The approaches include pre-processing of the facial samples to normalise any variation, before feeding it to the face recognition algorithm \citep{Shan2003Illumination,HussainShah2015Robust,Zou2007Illumination,Shinwari2019Comparative,ruiz2008illumination}. 

Databases have been introduced to facilitate the development of FR models with light, pose, expression and other variations. The Yale b, CMU Multi-PIE, AR, Postech01 and UHDB31 \citep{Gross2010Multi-pie,Martinez1998AR,Georghiades2001From,Dong2001Asian,Le2017UHDB31:} are a few of the datasets that have incorporate face variations and either focus on a single variation or a combination of different variations. Despite the developments of such datasets, the number of the variations and human subjects remain limited along with the fact that these datasets are usually acquired in controlled environments and therefore not being able to represent in-the-wild conditions. 

Relighting techniques have been introduced in the literature with impressive results and able to introduce lighting into the face images without the degradation of the image by artifacts \citep{Zhou2019Deep,Wang2020Single} giving the ability to augment in-the-wild face datasets. Thus, providing a solution to the limited variation of lighting and human subjects in face datasets, which is discussed in section \ref{effect of light}.

\section{Methodology}
In this section the various techniques and methodologies used in this work are detailed.   
\label{Methodology}
\subsection{Face Re-Lighting Method}
\label{relight}
In this work, the lighting variation is applied to the CelebA-HQ dataset using the Deep Single Image Portrait Relighting (DPR) technique \citep{Zhou2019Deep}. In this method a CNN is trained to generate a relighted image based on a Spherical Harmonics (SH) description of a lighting source. The method achieves SoA results, and in particular avoids introducing artifacts to the relighted samples - a drawback of other re-lighting methods that were considered for use in this study. The selected DPR method is trained on the well-known CelebA-HQ dataset which provides good variability in term of subject identity, combined with consistent face image quality. This makes the combination of the DPR re-lighting methodology and CelebA-HQ ideal for this work as  side-effects are eliminated due to either variable facial sample quality or re-lighting artifacts, either of which could distort our experimental outcomes. 

In this work we restricted our experiments to a select set of directional lighting components in order to gain a better understanding of the overall effect of directional lighting. The selected scenarios that are examined include lighting from 4 main directions: right, left, top and bottom of the face image. This has the added benefit of keeping the computation requirements for experiments bounded to reasonable time-frame, with most individual experiments completing in less than a 48 hour period. 

\begin{figure}[!t]
\centering
\includegraphics[scale=.5]{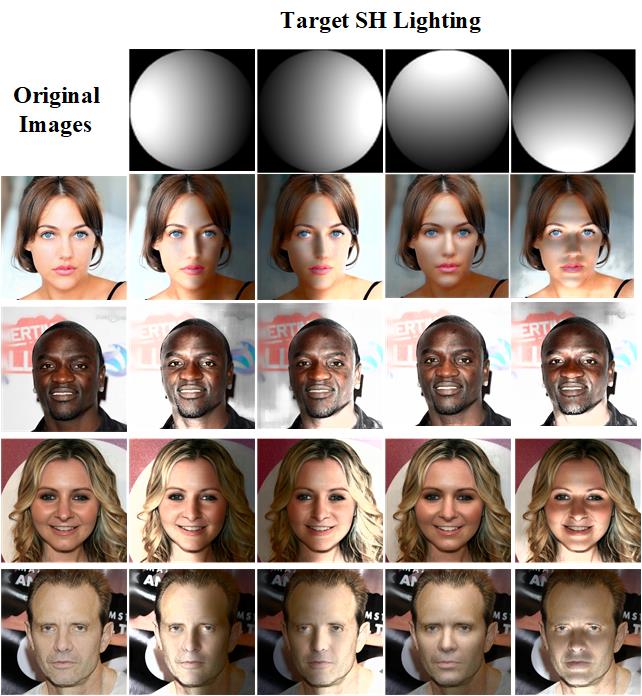}
\caption{The target SH lighting that is examined in this work is presented on the top row. The original images of CelebA-HQ are on the left column. Examples of lighting injected in the original images using the DPR method \citep{Zhou2019Deep} are shown for each examined illumination scenario (right, left, top \& bottom).}
\label{light examples}
\end{figure}

The representative Spherical Harmonic (SH) lighting sources used are shown in Fig.\ref{light examples}. More SH lighting scenarios can be found in \footnote{https://zhhoper.github.io/dpr.html \label{relight github}}. The illumination variations (right, left, top and bottom) are introduced to each sample of the CelebA-HQ dataset, resolving with 4 new sets of the CelebA-HQ, each containing of one illumination variation (CelebA-HQ-right, CelebA-HQ-left, CelebA-HQ-top and CelebA-HQ-bottom). Examples of the CelebA-HQ samples after introducing the illumination variations are illustrated in Fig.\ref{light examples}. 
It can be seen from Fig.\ref{light examples} that the DPR method has high quality outputs incorporating the target SH lighting to the images realistically and without generating any artifacts to the face images. Instructions on how to generate the sets of CelebA-HQ with the different illumination scenarios are given in the Github repository of this work \footnote{https://github.com/C3Imaging/Deep-Learning-Techniques/tree/Quantify-Retrain-FR-for-Light \label{our github}}.  

\subsection{Face Recognition Model}
\label{FR model}
A public reference implementation of the ArcFace \citep{Deng2019Arcface:} model is available, as the authors have released optimized, pre-trained, weights for the model.This reference ArcFace model has high performance on the dataset used in this work and provides a useful public baseline for future performance comparisons. This has motivated our use of ArcFace throughout this study. Other SoA FR models such as FaceNet \citep{Schroff2015Facenet:} or CosFace \citep{Wang2018Cosface:} do not provide reference implementations and thus restrict direct experimental comparisons. An unofficial, but public, implementation of FaceNet \footnote{https://github.com/davidsandberg/facenet \label{facenet david}} was also tested but could not provide a similar level of performance on the baseline or test datasets used in this work. 

In this work, the recommended workflow, by the authors of ArcFace is followed before the face samples are fed to the network. The MTCNN \citep{Xiang2017Joint} is used to detect and crop the face samples. The detected area is cropped and resized to $112 \times 112$, using bilinear interpolation, before passing to the ArcFace network which calculates the 512-embedding corresponding to a facial sample. Finally, the cosine similarity is used to compute the score representing the identity similarity when comparing two embeddings. The pretrained network used in this work is provided by the authors of ArcFace and can be found in \footnote{https://www.dropbox.com/s/ou8v3c307vyzawc/model-r50-arcface-ms1m-refine-v1.zip?dl=0 \label{arcface net}}.

Due to the introduction of the lighting variation and other factors, the face detection is not able to process all the face images from the CelebA-HQ sets. In the experiments only the images which the face detection network was able to process in all the illumination scenarios along with the original image, are used in order to keep the consistency in the experiments. Therefore, from the initial 30,000 images, in this work 28,222 are used from each set (CelebA-HQ-left, right, top, bottom and original). For the requirements of this study the dataset is divided into a train and test set with 19,570 images from 4k identities and 8,654 images from 2k identities, respectively. This is applied to each CelebA-HQ-set. A list of the images used in the experiments can be found in \textsuperscript{\ref{our github}}. 

\subsection{Using ROC Curves as a Metric}
\label{roc}
The ROC curve is created by plotting the true positive rate (TPR) against the false positive rate (FPR) at various threshold settings \citep{Mansfield2006Information}. More specifically, to compute an ROC curve, an equal number of positive-identity and negative-identity pairs are created. Using the corresponding embeddings (extracted from the FR model for each image) of the pairs, similarity scores are calculated and used to plot the ROC curve. The closer an ROC curve is to unity, the better the performance of the FR model on the selected samples. More information regarding the ROC and its interpretation and use can be found in \citep{varkarakis2020validating}. 

\section{Quantifying the Effects of Illumination on the Face Recognition}
\label{effect of light}
In this section, the experiments conducted to quantify the effect of different lighting conditions on the FR's performance are presented with a discussion of the findings.

\subsection{Experiments on Initial Lighting Variations}
\label{roc before}
\begin{figure}[!t]
\centering
\includegraphics[width=0.99\textwidth]{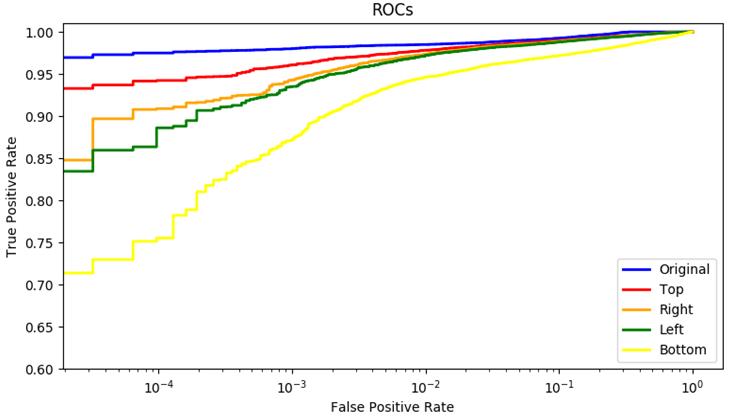}
\caption{ROC curves, representing the performance of the FR model \textsuperscript{\ref{arcface net}} on the original images and the 4 initial directional illumination scenarios (left, right, top, bottom) examined in this work.}
\label{before finetune}
\end{figure}

The effects of the 4 directional lighting scenarios shown in Fig.\ref{light examples} on the FR's performance are examined. Initially an ROC curve is calculated using only the samples from the test set of the original CelebA-HQ (ROC-Original) following the procedure described in section \ref{roc}. 
All possible positive pairs from the test set of the original CelebA-HQ are used, in total 31k image pairs and an equal number of negative pairs are created randomly. Using the corresponding FR embeddings, the similarity scores are calculated and used to plot the ROC curve. The ROCs corresponding to re-lighting augmented scenarios are calculated following a similar procedure. The same positive and negative identity pairs as in ROC-Original are used but one of the samples from each pair has a re-lighting augmentation applied. Thus resulting in 4 main ROC curves (ROC-Left, ROC-Right, ROC-Top, ROC-Bottom) representing the FR’s performance in each illumination scenario. The positive and negative pairs used to compute each ROC can be found in \textsuperscript{\ref{our github}}. The resulting ROCs enable a direct comparison of the effects of different types of directional illumination with the original set of test image pairs and between them. This is presented in Fig.\ref{before finetune}.\\

From Fig.\ref{before finetune} the initial experimental results are largely self consistent and show well-defined performance degradation of the FR which is largely consistent with what might be expected. The ROC-Original curve illustrates that the FR model has a SoA performance on the non-augmented test dataset approaching close to unity, of 0.99 TPR on the corresponding to $10^{-4}$ FPR value. 
The re-lighting augmented ROC curves show significant deviations from this baseline performance and are largely consistent with what might be expected. Thus, the smallest deviation is for the ROC-Top, which starts at 0.925 TPR, followed by the ROC-Right and ROC-Left curves at 0.86 and 0.85 respectively. The worst performing ROC is that of the bottom light, starting a TPR of only 0.725. Looking at the examples shown in Fig.\ref{light examples} these results make sense - the top lighting augmentation causes the least distortion to the facial image from a human perspective, whereas the bottom-lighting creates more obvious distortions in the facial features. Finally, the left/right lighting augmentations would be expected to have similar effects due to the symmetry of a human face. Note that the slight variation between ROC-Left and ROC-Right is most likely due to slight left-right pose variations in some facial samples leading to eccentricities in the corresponding lighting augmentations.       

\begin{figure}[!t]
\centering
\includegraphics[scale=.5]{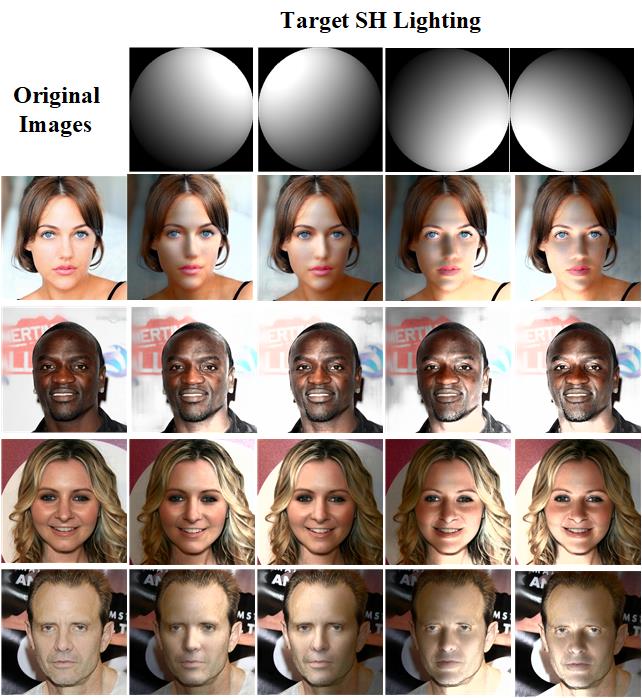}
\caption{The additional SH lighting that is examined in section \ref{roc before 2} is presented on the top row. The original images of CelebA-HQ are on the left column. Examples of lighting injected in the original images using the DPR method \citep{Zhou2019Deep} are shown for each examined illumination scenario (top-left, top-right, bottom-left, bottom-right).}
\label{light examples 2}
\end{figure}

\subsection{Experiments on Additional Lighting Variations}
\label{roc before 2}
The initial results shown in Fig.\ref{before finetune} encouraged a more extensive set of experiments to include additional top-right, top-left and bottom-right, bottom-left lighting augmentations, to further improve our understanding of mixed directional lighting modes. The new SH lighting used and examples of the CelebA-HQ samples after introducing these illumination variations are illustrated in Fig.\ref{light examples 2}. The goals of this additional set of experiments were to provide a second validation of our results, and in addition to explore the effects of more varied re-lighting augmentations. 

Due to the introduction of the new lighting variations, the face detection is not able to process all the face images from the test set of the CelebA-HQ. Similarly as in section \ref{FR model}, only the images which the face detection network was able to process in all 8 illumination scenarios and the original images are used in order to keep the consistency in the experiments. Therefore, the initial test set of 8,654 images from 2k identities is reduced to 8,552 images from 1,979 identities. 
In order to calculate the ROCs, corresponding to the original images and the 8 illuminations scenarios the procedure described in \ref{roc} and \ref{roc before}, is followed, using the new tests. As the size of the test set is reduced, so is the number of all possible positive image pairs used to compute the ROCs. For this set of experiments 30k positive pairs and an equal number of negative pairs are used. These pairs can be found in \textsuperscript{\ref{our github}}.\\

\begin{figure}[!t]
\centering
\includegraphics[width=0.99\textwidth]{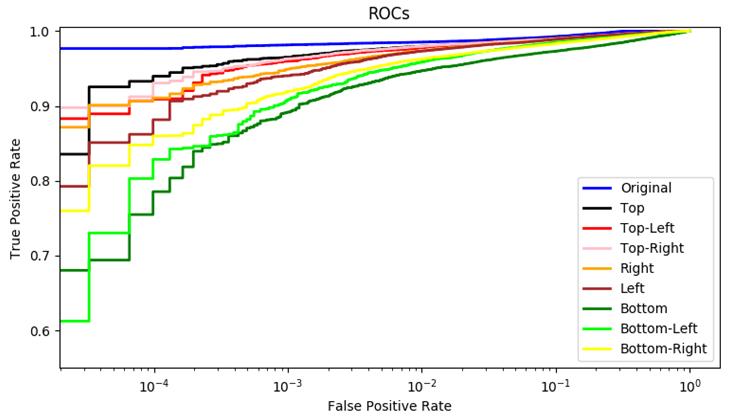}
\caption{Additional ROC curves, representing the performance of the FR model \textsuperscript{\ref{arcface net}} on the original images and the 8 directional illumination scenarios (left, top-left, bottom-left, right, top-right, bottom-right, top and bottom) examined in this work.}
\label{before finetune 2}
\end{figure}

The primary directional ROCs curves (top, bottom, left \& right) presented in Fig.\ref{before finetune 2}, differ slightly from those of Fig.\ref{before finetune}, as the image pairs used in these experiments are different. However their behaviour has a broadly similar characteristic. The bottom, bottom-left and bottom-right lighting augmentations are seen to be the most challenging for the FR task, while the top-left and top-right illuminations have the least effect on the FR's performance. There is a small drop and a small increase on the FR's performance from the top and right light respectively, compared to the Fig.\ref{before finetune}. These are attributed to the use of different pairs. These results are useful as a baseline for the section \ref{finetune sec}, as they help demonstrate that fine-tuning on the primary set of directional lighting augmentations can generalize across a broader range of directional lighting effects.  

\subsection{Experiments on Public Face Illumination Data-sets}
\label{experiments on real datasets}
Similar experiments as in \ref{roc before} and \ref{roc before 2}, were conducted using face datasets which include illumination variation in their samples, to measure the effect of their lighting scenarios on the FR's performance. More specifically the AR and Postech01 \citep{Martinez1998AR,Dong2001Asian} face datasets were used. These experiments did not show any measurable degradation of the FR's performance across the illumination conditions provided in these data-sets. This is attributed to the relatively limited variation of the illumination scenarios, the numbers of human subjects, the total number of images and the controlled environments in which these face datasets were acquired. This is also indicated through the information regarding these datasets, provided in Table \ref{tab:table1}. Thus, revealing the need for the development of a face dataset that resembles in-the-wild conditions with illumination variation, which is further discussed in section \ref{conclusion and future work}. 

\begin{table}[!t]
\centering
 \caption{Summary Information of Face Data-sets with Illumination Variations. (The number of Pose Variations in this table is referred with regards to the Illumination Variation)}
    \label{tab:table1}
    \begin{tabular}{|p{2.5cm}|p{0.7cm}|p{1.9cm}|p{1.9cm}|}
 \hline 
Face Data-sets & \#IDs & \#Illumination Variations & \#Pose  Variations \\
 \hline
 AR \citep{Martinez1998AR}  & 126  & 3 & 1 \\
\hline
Postech01 \citep{Dong2001Asian}   & 103  & 4 & 1 \\
\hline
    \end{tabular}
\end{table}

\section{Fine-Tuning the Face Recognition model on Directional Illuminations} 
\label{finetune sec}
In the first part of this work the potential effects of in-the-wild lighting conditions, in particular directional lighting effects on a SoA neural face recognition method have been demonstrated and quantified. The next step is to determine whether the FR can be tuned to compensate for these effects. In this section the selected FR model is fine-tuned, using a similar approach to \citep{varkarakis2020deep} with samples augmented with directional lighting.  

\subsection{Fine-tuning Process \& ROCs Computation}
\label{finetune method}

The initial pretrained network provided by the authors of Arcface \textsuperscript{\ref{arcface net}}, is fine-tuned using a training set comprising samples from the original CelebA-HQ dataset and samples with all 4 primary directional lighting augmentations (CelebA-HQ-Left, Right, Top and Bottom). In total 97,850 high-quality facial samples were used for fine-tuning, or 19,750 from the original data and each of the four primary lighting sub-category.

For the re-training process the standard Arcface loss function is used, with the learning rate set to 0.005 and a batch size of 128, following the instructions from the authors of ArcFace. The network is fine-tuned for 40 epochs, as the number of images used is relatively large and all network layers are unfrozen for the fine-tuning process. After 40 epochs the network showed satisfactory results on the training data used and therefore stopped. Longer training could result in over-fitting to the training data and thus not being able to generalise. More details regarding the fine-tuning process and the corresponding training code can be found at \footnote{https://github.com/deepinsight/insightface \label{arcface github}}. The fine-tuned network resulting from this re-training process is released at \textsuperscript{\ref{our github}}. 

The fine-tuned network, is used to calculate the embeddings of the test samples. The same procedure as described in \ref{roc} and \ref{roc before} is followed to calculate the ROC-Original-finetuned (ROC-Original-FT) and the ROCs corresponding to the 8 different illumination scenarios (ROC-Left-FT, Right-FT, Top-FT, Bottom-FT, Top-Left-FT, Top-Right-FT, Bottom-Left-FT, Bottom-Right-FT,), using the positive and negative pairs from section \ref{roc before 2}. These ROCs are compared with the ROC-Original-FT and between them to explore whether the fine-tuned FR model is able to handle the variation in illumination as well as whether can generalise across the illuminations that were not used for the fine-tuning process. 

\begin{figure}[!t]
\includegraphics[width=0.99\textwidth]{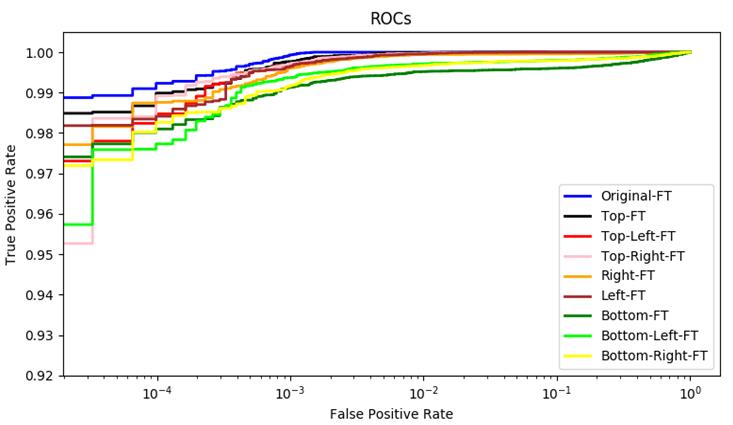}
\caption{ROC curves, representing the performance of the finetuned FR network \textsuperscript{\ref{our github}} on the original images and the 8 directional illumination scenarios (left, top-left, bottom-left, right, top-right, bottom-right, top and bottom) examined in this work. }
\label{after finetune}
\end{figure}

\subsection{ROCs Comparison}
\label{finetune results}
The ROCs representing the performance of the fine-tuned FR model on the original images and on the 8 directional lightings are presented in the Fig.\ref{after finetune}.
From Fig.\ref{after finetune} it is illustrated that the ROCs corresponding to the fine-tuned FR model on the 4 main illuminations (left, right, top, bottom) used in the fine-tuning process (Fig.\ref{after finetune}) are at higher levels compared to the ROCs corresponding to the performance of the initial network (Fig.\ref{before finetune 2}) on these illumination scenarios. More importantly, the ROCs corresponding to the fine-tuned model on the 4 illuminations scenarios that are not used during the fine-tuning (top-left, top-right, bottom-left, bottom-right), are also at higher levels, thus showing that the network is able to generalise to other variations of illumination that it was not trained on. Overall, the performance of the fine-tuned FR model on any given illumination scenario has increased and its above 0.95 TPR, even on the lower FPR values. Notably, the ROCs are very close to the performance of the FR on the original images. Therefore, concluding that the FR model when trained with lighting variation is able to adopt and handle face samples that include illumination and achieve high accuracy results and also generalise across different illumination variations that are not used during fine-tuning. Thus, showing that the illumination can be compensated through training methods and augmentation techniques eliminating the need for pre-processing methods correcting the lighting, which are not optimal for use in neural accelerators.

\section{Conclusion \& Future Work}
\label{conclusion and future work}

It is clear from the results of the experiments, illustrated in Fig. \ref{before finetune} \& \ref{before finetune 2} that fully end-to-end neural FR solutions will be challenged by in-the-wild lighting conditions. As was indicated in \citep{mehdipour2016comprehensive} this problem is typically solved by additional pre-processing of image samples to correct for lighting conditions. 
In section \ref{finetune results} the practicality of fine-tuning a high-performing neural FR model has been demonstrated, recovering performance levels close to the original baseline for such lighting conditions. The fine-turning process also indicated that generalization from the primary directions to combinations of directional lighting is achieved - a promising result given the non-linear nature of lighting conditions. Note that providing a broader and more varied range of re-lighting samples and refining the training methodology to identify the more sensitive network layers in the ArcFace model should further improve these results, but even as they stand it is clear that a full end-to-end neural FR can be realized.\\

These initial results, especially the effectiveness of the fine-tuning process are very promising. They suggest that SoA neural FR algorithms can be fine-tuned to handle difficult in-the-wild acquisition conditions such as directional lighting. However there are other challenges for FR algorithms in-the-wild, including those listed in the introduction. A broader study on factors that can affect FR is indicated. In this regard the availability of several large 3D facial model datasets \citep{yang2020facescape} could provide sufficient individual identities and support more complex data variations to support such a study.

\section*{Acknowledgments}
This research is funded by (i) the Science Foundation Ireland Strategic Partnership Program (Project ID: 13/SPP/I2868), (ii) Irish Research Council Enterprise Partnership Ph.D. Scheme (Project ID: EPSPG/2020/40) and, (iii) Xperi Corporation, Ireland.  

\bibliographystyle{model2-names}
\bibliography{manuscript}

\end{document}